\def\year{2019}\relax
\documentclass[letterpaper]{article} 
\usepackage{aaai19}  
\usepackage{times}  
\usepackage{helvet}  
\usepackage{courier}  
\usepackage{url}  
\usepackage{graphicx}  
\usepackage{eso-pic}

\usepackage{amsmath}
\usepackage{lipsum}	
\usepackage{algorithm}
\usepackage[noend]{algpseudocode}
\usepackage[square, comma, sort&compress, numbers]{natbib}

\def\BState{\State\hskip-\ALG@thistlm}

\usepackage{comment}
\usepackage{color}
\usepackage{etoolbox}
\newbool{inccomment}
\setbool{inccomment}{true}
\newcommand{\xc}[1]{\ifbool{inccomment}{{\color{blue} #1}}{}}
\newcommand{\yc}[1]{\ifbool{inccomment}{{\color{red} #1}}{}}

\usepackage{graphicx}
\usepackage{subcaption}

\begin{document}
\title{2PFPCE: Two-Phase Filter Pruning Based on Conditional Entropy}
\author{Chuhan Min\thanks{Chuhan Min and Aosen Wang contribute equally to this work.}\thanks{This work was done during the internship of Chuhan Min and Aosen Wang at Midea Corporate Research Center.}\\
University of Pittsburgh\\
chm114@pitt.edu\\
\And
Aosen Wang\footnotemark[1]\footnotemark[2]\\
SUNY at Buffalo\\
aosenwan@buffalo.edu\\
\AND
Yiran Chen\\
Duke University\\
yiran.chen@duke.edu\\
\And
Wenyao Xu\\
SUNY at Buffalo\\
wenyaoxu@buffalo.edu\\
\And
Xin Chen\thanks{Corresponding author}\\
Midea Corporate Research Center\\
chen1.xin@midea.com\\
}
\maketitle

\begin{abstract}
Deep Convolutional Neural Networks~(CNNs) offer remarkable performance of classifications and regressions in many high-dimensional problems and have been widely utilized in real-word cognitive applications.
However, high computational cost of CNNs greatly hinder their deployment in resource-constrained applications, real-time systems and edge computing platforms.
To overcome this challenge, we propose a novel filter-pruning framework, two-phase filter pruning based on conditional entropy, namely \textit{2PFPCE}, to compress the CNN models and reduce the inference time with marginal performance degradation.
In our proposed method, we formulate filter pruning process as an optimization problem and propose a novel filter selection criteria measured by conditional entropy.
Based on the assumption that the representation of neurons shall be evenly distributed, we also develop a maximum-entropy filter freeze technique that can reduce over fitting. 
Two filter pruning strategies -- global and layer-wise strategies, are compared. 
Our experiment result shows that combining these two strategies can achieve a higher neural network compression ratio than applying only one of them under the same accuracy drop threshold.
Two-phase pruning, that is, combining both global and layer-wise strategies, achieves $\sim 10\times$ FLOPs reduction and 46\% inference time reduction on VGG-16, with 2\% accuracy drop.
\end{abstract}

\section{Introduction}
\label{sec:introduction}



Deep Convolutional Neural Networks~(CNNs) have been widely utilized in many applications and achieved remarkable success in computer vision~\cite{szegedy2015going}, speech recognition~\cite{abdel2012applying}, natural language processing~\cite{collobert2008unified}, etc.
Going deeper has been proven as an effective approach to improve the model accuracy in solving high-dimensional problems~\cite{ba2014deep,szegedy2015going}.
However, when the network depth increases, the number of parameters of the neural network increases too.\\
\begin{figure}[!t]
\centering
\includegraphics[width=1.0\columnwidth]{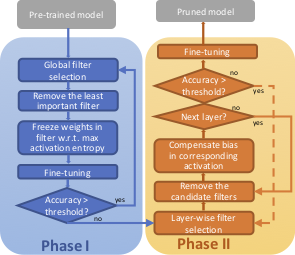}
\caption{Basic scheme of our proposed 2PFPCE. This scheme show the network pruning process consisting phase I and II.}
\vspace{-10pt}
\label{fig:4_exp_method}
\end{figure}
Model compression techniques aim at
reducing the storage and computational costs of deep neural networks~\cite{lecun1990optimal, gong2014compressing, han2015deep, wen2016learning}.
Network pruning is one important example of model compression techniques that can reduce the network complexity and suppress the over-fitting issue simultaneously.
Han et al.~\cite{han2015deep, han2015learning} proposed to reduce network parameters by pruning the weights with small magnitudes and then retrain the network in an iterative manner to maintain the overall accuracy.
Majority of the pruned parameters is actually from fully connected layers.
Since fully connected layers contribute to very small portion of the total floating point operations (FLOPS), e.g., less than 1\% in VGG-16~\cite{simonyan2014very}, the overall computational cost reduction achieved by this method is very limited~\cite{wen2016learning}. 
Moreover, the random distribution of the removed weights in memory hierarchy also incurs a higher cache miss rate, which greatly harms the actual performance acceleration obtained in real systems~\cite{wen2016learning}. 
In some recent work on CNNs~\cite{szegedy2015going, he2016deep}, the fully connected layers are replaced by average pooling layers in order to build a deep architecture with hundreds of layers. 
Recently, more and more work focused on pruning convolutional layers to reduce computational cost in inference time~\cite{li2016pruning,molchanov2016pruning,luo2017entropy}.
Despite the significant weight sparsity in fully connected layers, the non-structured random connectivity ignores cache and memory access issues as indicated in~\cite{wen2016learning}.
In some recent work on CNNs~\cite{szegedy2015going, he2016deep}, the fully connected layers are replaced by average pooling layers in order to build a deep architecture with hundreds of layers.
The computational cost of the convolutional layers, hence, dominates when the networks become deeper.
We note that CNNs with a large scale usually have significant redundancy of their filters and feature channels, which offer a large compression and pruning space.


In this work, we propose a \textbf{T}wo-\textbf{P}hase \textbf{F}ilter \textbf{P}runing framework based on \textbf{C}onditional \textbf{E}ntropy, referred to as \textit{2PFPCE}, to prune the filters of CNNs based on conditional entropy in a two-phase manner, as shown in Figure~\ref{fig:4_exp_method}. 
The key idea of our proposed approach is to establish a quantitative connection between the filters and the model accuracy. 
We adopt global pruning as phase-one and layer-wise pruning as phase-two.
In Phase I, the filters with the minimum conditional entropy is pruned filter-by-filter, followed by an iterative fine-tuning constrained by an accuracy drop threshold.
In Phase II, the filters are pruned layer-by-layer in a greedy manner based on conditional entropy, followed by also a fine-tuning of the neural network constrained by the accuracy loss threshold.
Our major contributions can be summarized as:
\begin{itemize}
	\item We calculate the conditional entropy over the filters in a convolutional layer, i.e. the distribution of entropy conditioned on the network loss. We also propose to use conditional entropy as a criteria to select the filters to be pruned in our method:
    \item Based on the assumption that the information of the neurons in a layer shall be uniformly distributed, we propose a novel fine-tuning approach where the weights in a filter corresponding to the neuron with the maximum entropy is kept constant during the back-propagation to reduce over-fitting.
    \item Based on our observation on the different pruning efficiencies of global and layer-wise pruning strategies, we propose to combine these two strategies to achieve a higher compression ratio of the neural network compared to applying only one strategy in network pruning.
Experimental results show that 2PFPCE can achieve a reduction of 88\% filters on VGG16 with only 2\% accuracy degradation. The data volume is decreased from 310784 bytes to 49165 bytes and the inference time is $\sim 54\%$ of the original model. 
\end{itemize}

\section{Related works}
\label{sec:preliminary}



\subsection{Model compression}
The compression techniques of convolutional layers can be roughly categorized into the following three types according to their approximation level:

\textbf{Pruning} reduces the redundancy in parameters which are not sensitive to the performance at a level of weight and filter.
Network pruning, which aims at reducing the connectivities of the network, is a classic topic in model compression and has been actively studied in the past years. 
Pruning has been performed at weight level~\cite{han2015deep, wen2016learning} and filter level~\cite{li2016pruning, molchanov2016pruning}.
\textbf{Quantization} compresses the network by reducing the number of bits required to represent the weights~\cite{han2015deep}.
Binarization~\cite{rastegari2016xnor} is an extreme case of quantization where each weight is represented using only 1-bit.

\textbf{Convolution reconstruction} divides convolution into subproblems based on organization of filters at layer level. 
Low rank approximation~\cite{denil2013predicting, zhang2015efficient, tai2015convolutional, ioannou2015training} imitate convolutional operations by decomposing the weight matrix as a low rank product of two smaller matrices without changing the original number of filters.
Based on the correlation between groups of filters, ~\cite{cohen2016group, zhai2016doubly} build a convolutional layer from a group of base filters.
FFT convolution~\cite{vasilache2014fast} designs a set of leaf filters with well-tuned in-register performance and reduce convolution to a combination of these filters by data and loop tiling. 

\textbf{Knowledge distillation}~\cite{hinton2015distilling} compresses an ensemble of deep networks (teacher) into a student network with similar depth by applying a softened penalty of the teacher’s output to the student.\\
This compression method works at network level.

There is no golden rule to measure which one of the three kinds of approach is the best.
In this work, we focus on filter pruning.
There exist some heuristic criteria to evaluate the importance of each filter in the literature such as APoZ (Average Percentage of Zeros)~\cite{hu2016network}, $\ell_1$-norm~\cite{li2016pruning} and Taylor expansion~\cite{molchanov2016pruning}.
\begin{itemize}
	\item APoZ (Average Percentage of Zeros)~\cite{hu2016network}: calculates the sparsity of each channel in output feature map as its importance score $\frac{\sum_{k}^{N}\sum_{j}^{M}f\left ( O_{c, j}(k)=0\right )}{N\times M}$.
    \item $\ell_1$-norm~\cite{li2016pruning}: measure the relative importance of a filter in each layer by calculating the sum of its absolute weights $\sum\left | \textit{F}_{i, j} \right |$, i.e., its $\ell_1$-norm $\left \| F_{i, j} \right \|_{1}$.
    \item Taylor expansion~\cite{molchanov2016pruning}: approximate change in the loss function with accumulation of the product of the activation and the gradient of the cost function w.r.t. to the activation $\left | \frac{1}{M}\sum_{m}\frac{\delta C}{\delta z_{l, m}^{(k)}}z_{l, m}^{(k)}  \right |$.
\end{itemize}

Unlike above mentioned criterion, we directly quantize contribution of each filter to accuracy via conditional entropy, discussed following section.

\subsection{Information Plane}
There is a growing interest in networking understanding and this motivates our information guided pruning.
~\cite{tishby2015deep} proposed to analyze DNNs in the \textit{Information Plane}.
The goal of the network is to optimize the Information Bottleneck (IB) trade-off between compression and prediction, successively, for each layer.

Two properties of the IB are very important in the context of network pruning.
The first is the necessity of redundancy during model training.
According to~\cite{tishby2015deep}, the Stochastic Gradient Decent (SGD) optimization has two different and distinct phases: empirical error minimization (ERM) and representation compression.
In ERM, redundancy is necessary since the high non-convex optimization is hard to be solved with current technologies.
Considering convergence rate, reducing model size after its training is more time efficient.
The second is the conditional distribution of output $y$ on $\widetilde{x}$, i.e. $p\left ( y \mid \widetilde{x} \right )$, where $\widetilde{x}$ is the compact expression of input $x$ follows from the Markov chain condition $Y\leftarrow X\leftarrow \widetilde{X}$.
It is important to notice that this not a modeling assumption and the quantization $\widetilde{x}$ is not used as a hidden variable in a model of the data. 
Hence, a network can be decomposed to a cascade of subnetworks with its compact input feature maps as input and original model's output as output.

On information plane, Mutual Information (MI) quantifies the average number of relevant bits that the input variable X contains about the label Y.
\begin{equation}
\begin{split}
I\left ( X, Y \right ) & =\sum_{(x, y)\in A}p(x, y)\log[\frac{p(x,y)}{p(x)p(y)}] \\
& =\sum_{(x, y)\in A}p(x, y)\log[\frac{p(x \mid y)}{p(x)}] \\
& = H(X) - H(X|Y)
\end{split}
\end{equation}
The connection between mutual information and minimal sufficient statistics is based on its invariance to invertible transformations:
\begin{equation}
I(X, Y) = I(\psi (X), \phi (Y))
\end{equation}
for any invertible functions $\psi$ and $\phi$.
The invariance of the information measures to invertible transformations comes with a high cost.
For deterministic functions, the mutual information is insensitive to the complexity of the function or the class of functions it comes from~\cite{shwartz2017opening}.
If we have no information on the structure or topology of X, there is no way to distinguish low complexity classes from highly complex classes by the mutual information alone.

In this paper, instead of utilizing noise insensitive MI criteria, we propose to adopt conditional entropy in terms of error probability in guessing a finitely-valued random variable $X$ given another random variable $Y$. 

\section{Conditional entropy based compression}
\label{sec:method}


In this section, we first formulate compression as an optimization problem, then propose a conditional entropy based filter selection criteria and compare the statistical result of CIFAR10 on pre-trained VGG-16 model. 
Furthermore, we discuss the relationship of error probability and conditional entropy.

\begin{figure*}[!ht]
\centering
\begin{subfigure}{.7\columnwidth}
	\includegraphics[width=\columnwidth]{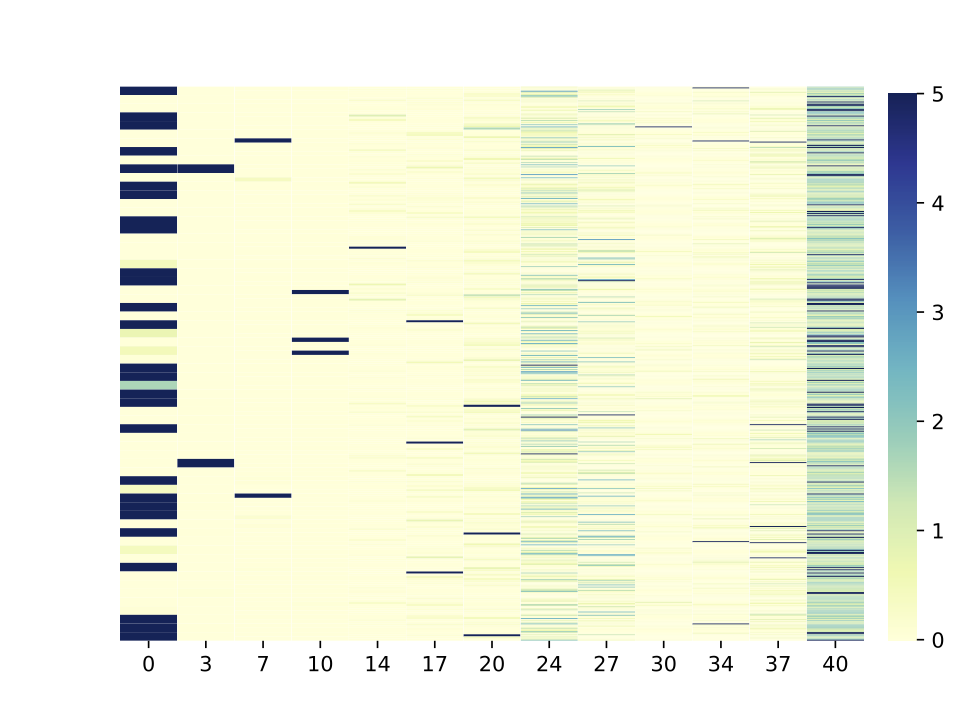}%
	\caption{Maximum conditional entropy.}%
	\label{maxv}%
\end{subfigure}\hfill%
\begin{subfigure}{.7\columnwidth}
	\includegraphics[width=\columnwidth]{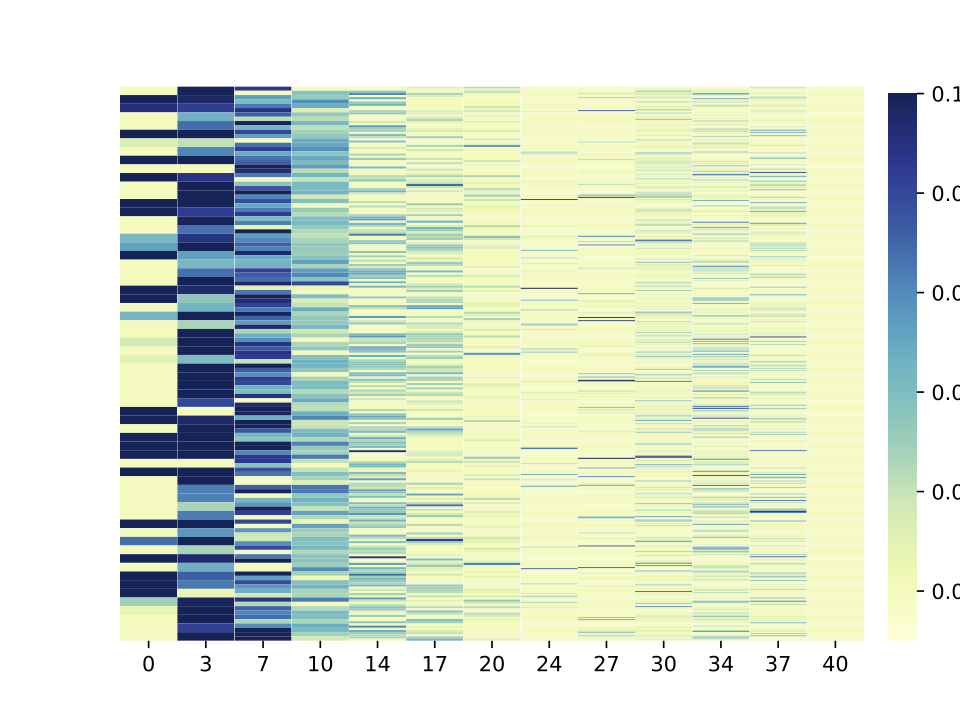}%
	\caption{Activation of max conditional entropy.}%
	\label{maxp}%
\end{subfigure}\hfill%
\begin{subfigure}{.7\columnwidth}
	\includegraphics[width=\columnwidth]{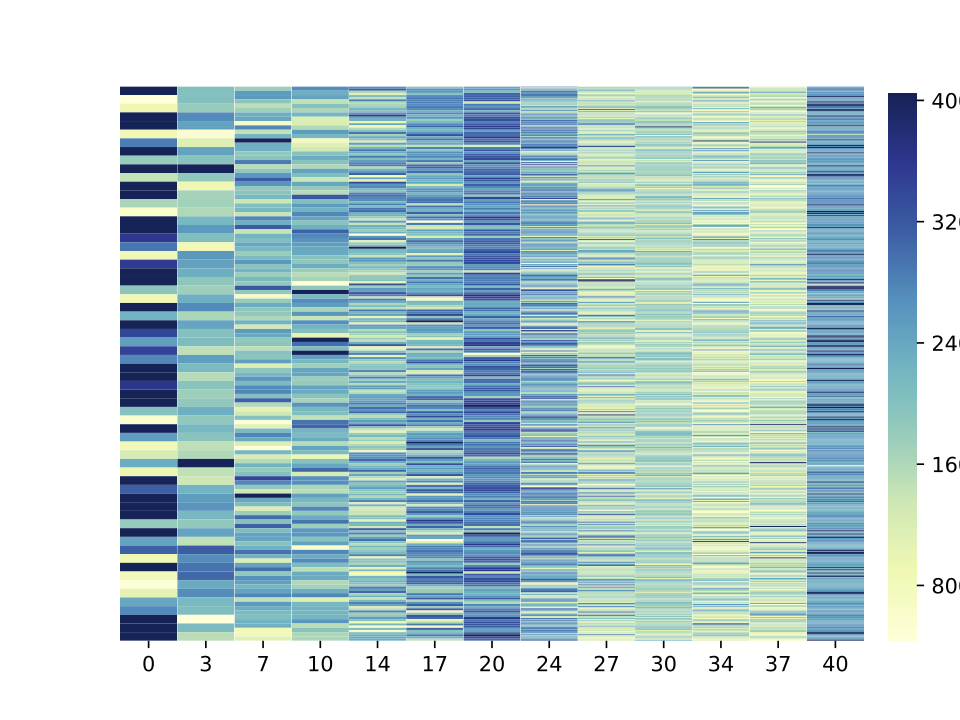}%
	\caption{Number of zero activations.}%
	\label{act_zero}%
\end{subfigure}\hfill%
\begin{subfigure}{.7\columnwidth}
	\includegraphics[width=\columnwidth]{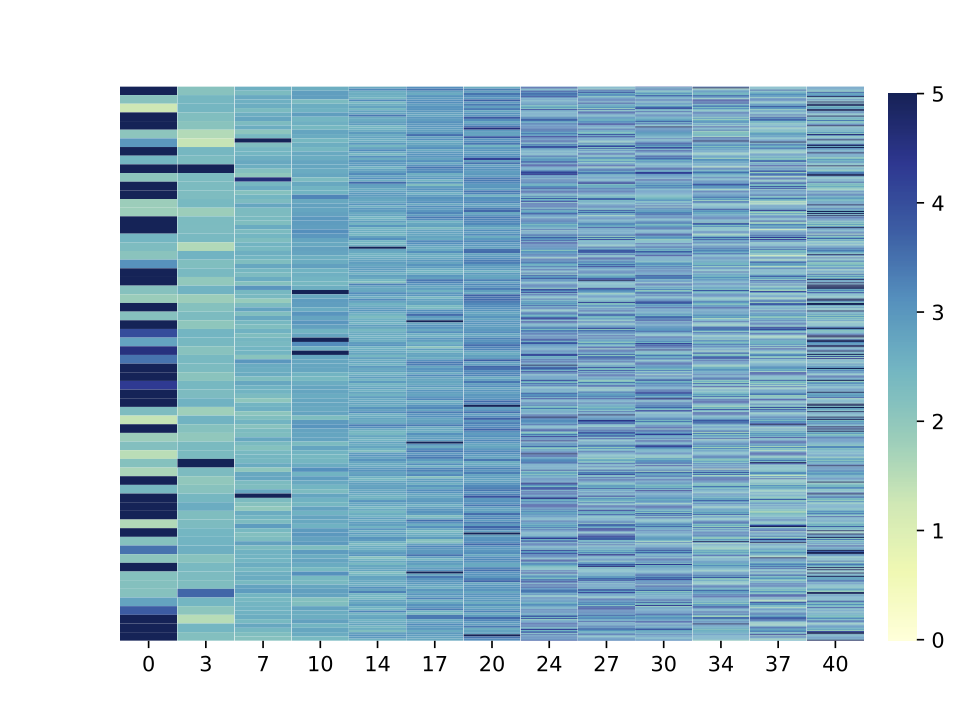}%
	\caption{Average conditional entropy of filter.}%
	\label{ent}%
\end{subfigure}\hfill%
\begin{subfigure}{.7\columnwidth}
	\includegraphics[width=\columnwidth]{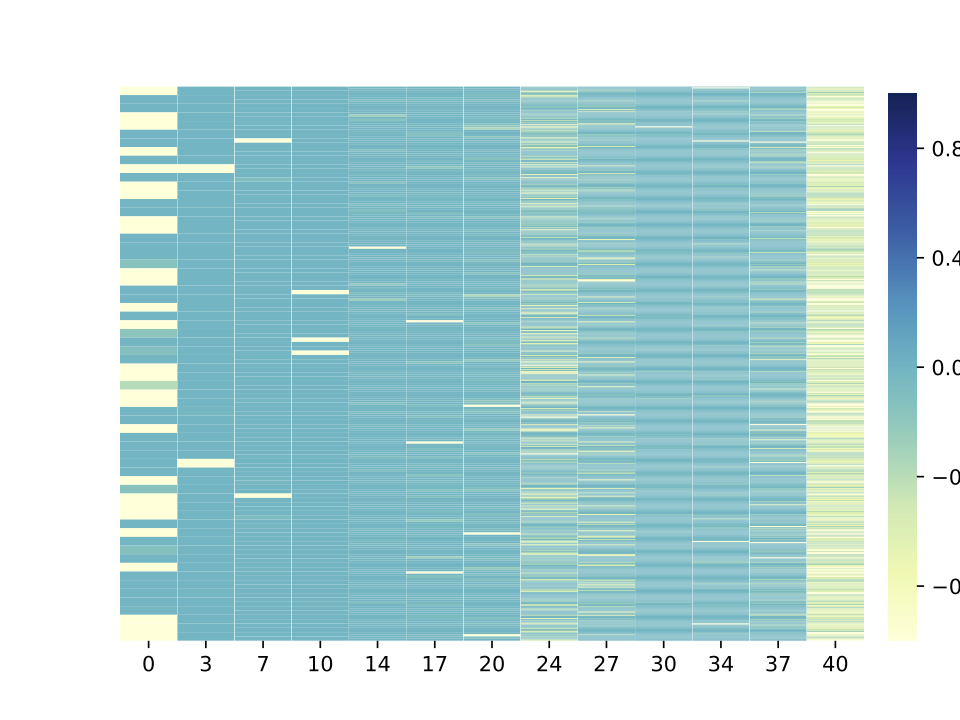}%
	\caption{Maximum conditional entropy filter ratio.}%
	\label{maxv_rate}%
\end{subfigure}\hfill%
\begin{subfigure}{.7\columnwidth}
	\includegraphics[width=\columnwidth]{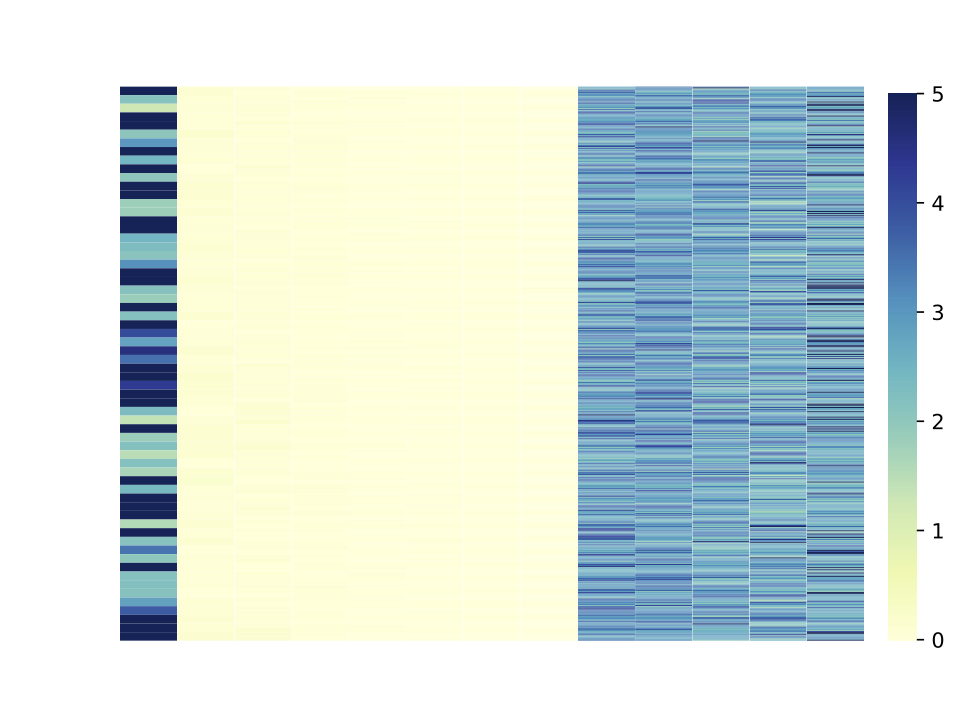}%
	\caption{Zero activation ratio.}%
	\label{act_zero_rate}%
\end{subfigure}%
\caption{Statistical result of VGG-16 on CIFAR10}
\label{fig:3_stat_vgg}
\end{figure*}

\subsection{Problem formulation}
In~\cite{shwartz2017opening,michael2018on}, each layer is seen as a \emph{single} random variable.
And the distribution is calculated by joining all neuron outputs in this layer.
However the same method might not be suitable for convolutional network, for losing the structure feature on both the input and the feature map.

Therefore we adopt the procedure to estimate the mutual information of each convolutional layer as in \cite{kraskov2004estimating,kolchinsky2017estimating}.
The first step is to use the entropy of each layer's output as the measurement of the information flow. 
The \textit{activation entropy} can be calculated using the function below where $p_i$ denotes the probability of \textit{i}-th filter in the feature map.
\begin{equation}
    H_{C_n} = \sum_{i=0}^{n} {p_i * \log p_i}
\end{equation}

Considering the general scenario of a neural network whose operation is parametrized by a vector~$\theta \in {\Re}^{W}$ (representing the weights), and whose input/output characteristics are described by a conditional probability distribution $p_{\theta }(x_{out} \mid x_{in})$, in which~$x_{in} \in {\Re}^{N}$ and~$x_{out} \in {\Re}^{M}$ denote input and output vectors, respectively.
The performance of this network on a given input is measured some loss function~$\varepsilon(x_{in}, x_{out})$.
If the probability of an input~$x_{in}$ to be encountered is defined as~$p(x_{in})$, the global error made by a network with parameter is given by
\begin{equation}
	E_{\theta }={\sum_{}^{x_{in}}}{\sum_{}^{x_{out}}}\epsilon (x_{in}, x_{out})p_{\theta }(x_{out}\mid x_{in})p(x_{in})
\end{equation}
We define $p_{\theta}(x) = p_{\theta}(x_{out}\mid x_{in})p(x_{in})$, $x = (x_{in}, x_{out}) \in {\Re}^{N+M}$.
It now combines both the parametrized properties of the network and the likelihood of input data.

When recognizing neural network as a stack of sub networks, the above definition holds true for each layer and thus error in convolutional layers: 
\begin{equation}
	E_{\theta }^{C}={\sum_{}^{C_{in}}}{\sum_{}^{x_{out}}}\epsilon (C_{in}, x_{out})p_{\theta }(x_{out}\mid C_{in})p(C_{in})
\end{equation}
Where~$C_{in}$ denotes the input feature maps of the convolutional layer.  
In this way, we reorganize the compression problem as a optimization problem and minimize the distance~$\left \|E_{\theta}^{C} - E_{{\theta}'}^{{C}'}\right \|$, where~${C}'_{in}$ is a minimum subset of $C_{in}$.

We adopt a d-dimensional binary vector $\sigma$: a 1 indicating the filter is selected, a 0 indicating the filter is discarded.
Notation $x_{\sigma}$ indicates the vector of selected features, that is, the full vector $x$ projected onto the dimensions specified by $\sigma$. 
Notation $x_{\widetilde{\sigma}}$ is the complement, that is, the unselected features.
The full feature vector can therefore be expressed as $x = \left \{ x_{\sigma},  x_{\widetilde{\sigma}} \right \}$. 
As mentioned, we assume the process $p$ is defined by a subset of the features, so for some unknown optimal vector $\sigma^{*}$, $p\left ( y \mid x \right )= p\left( y \mid x_{\sigma^{*}}\right)$.
We approximate $p$ using an hypothetical predictive model $q$, with two layers of parameters: $\sigma$ representing which filters are selected and $\tau$ representing the parameters used to predict y. 
Our problem statement is to identify the minimal subset of features such that we maximize the conditional likelihood of the training labels, with respect to these parameters.
For i.i.d data $D = \left \{ \left ( x^{i}, y^{i} \right );i = 1..N \right \}$ the conditional likelihood of the labels given parameters $\left \{ \sigma , \tau  \right \}$ is
\begin{equation}
	L\left(\sigma,\tau\mid D\right) = \prod_{i=1}^{N}q\left( y^{i}\mid x_{\sigma}^{i},\tau\right)
\end{equation}
The (scaled) conditional log-likelihood is
\begin{equation}
	l=\frac{1}{N} \sum_{i=1}^{N}\textup{log}q\left ( y^{i}\mid x_{\sigma}^{i}, \tau \right )
\end{equation}
This is the error function we wish to optimize with respect to the parameters $\left \{ \sigma , \tau  \right \}$; the scaling term has no effect on the optima, but simplifies exposition later.
We now introduce the quantity $p\left ( y \mid x_{\sigma} \right )$: this is the true distribution of the loss given the selected filters $x_{\sigma}$.
Multiplying and dividing $q$ by $p\left ( y \mid x_{\sigma} \right )$, we can re-write the above as,
\begin{equation}
l=\frac{1}{N} \sum_{i=1}^{N}\log\frac{q\left ( y^{i}\mid x_{\sigma}^{i}, \tau \right )}{p\left ( y^{i}\mid x_{\sigma}^{i}\right )}+\frac{1}{N} \sum_{i=1}^{N}\log{p\left ( y^{i}\mid x_{\sigma}^{i}\right )}
\end{equation}
The second term in (8) can be similarly expanded, introducing the probability $p\left ( y \mid x \right )$.
These are finite sample approximations, drawing data points i.i.d. with respect to the distribution.
We use $E_{xy}\left \{ . \right \}$ to denote statistical expectation, and for convenience we negate the above, turning our maximization problem into a minimization. This gives us
\begin{equation}
\begin{split}
-l & \approx E_{xy}\left \{ \log\frac{p\left ( y\mid x_{\sigma}\right )}{q\left ( y\mid x_{\sigma}, \tau \right )} \right \} + E_{xy}\left \{ \log\frac{p\left ( y\mid x\right )}{p\left ( y\mid x_{\sigma} \right )} \right \} \\
& - E_{xy}\left \{ \log{p\left ( y\mid x\right )} \right \}
\end{split}
\end{equation}

In the experiments below, we use the training loss as the single variable.
Then our problem statement is to identify the minimal subset of features such that we~\textit{maximize the conditional likelihood of the training loss, with respect to these parameters}.

\subsection{Filter selection algorithm}
The algorithm is illustrated below as shown in Algorithm~\ref{algo:3_procedure}.
For each sample, we calculate the cross entropy loss and output activation corresponding to each filter.
To achieve discrete statistical requirement, each parameter is multiplied by a factor of $1e4$ and quantized as 32-bit integer.
For each filter, $c\_val$ denotes 1-D distribution on output activation and $c\_bins$ denote a 2-D statistics on output activation conditioned on loss.

$c\_total$ is a number of activations per filter, i.e. samples in a dataset.
$act\_ent$ denotes the entropy of feature map, provided the distribution of output activation across the dataset. 
Notice here zero activations are excluded because it's considered to contain no information with respect to the next layer.
Given the probability of a specific output activation, $ent_i$ denotes the entropy of output activation conditioned on the distribution of cross entropy loss.
$con\_ent$ denotes the conditional entropy of a filter, which is an accumulation of $ent_i$.
Then, the $con\_ent$ is sorted in ascending and the filters corresponding to the top-$r$ $con\_ent$ are selected to be removed.
In above single-layer illustration, the layer to prune is predefined.
This can be generalized to multiple layers or the whole model.

\begin{algorithm}
\caption{Filter selection algorithm.}
\label{algo:3_procedure}
\hspace*{\algorithmicindent} \\\textbf{Input}: a baseline model $\boldsymbol{M}$, convolutional layer to prune $l$, training dataset $\textbf{x}_{train}$, number of filters to prune $r$
\hspace*{\algorithmicindent} \\\textbf{Output}: candidate filter(s) to prune $\sigma^r$
\begin{algorithmic}
\Procedure{ConditionalEntropy\_Calculation}{}
\State $ eps\_h = 1e4 $
\State $\textit{criteria} = CrossEntropy(reduce=false)$.
\For {$batch\_idx$ in $batches$} 
\State $output \gets \boldsymbol{M}(\textbf{x}_{train}[batch\_idx])$.
\State $loss[batch\_size] \gets \textit{criteria}(output, target)$.
\State $j=eps\_h * loss[batch\_size]$.
	\For {$k$ in $l$}
    \State $i=eps\_h * fmap\_out^{l}[k]$.
    \State $ans[k].extend(i, j)$.
    \State $c\_bins[i][j] += 1 $.
	\State $c\_val[i] += 1 $.
    \EndFor
\EndFor
\State $c\_total = \sum\limits_{i}^{c\_val}{c\_val[i]}$.
\State $act\_ent = \sum\limits_{j \neq 0}^{c\_val}{-\frac{c\_val[j]}{c\_total}*\log\frac{c\_val[j]}{c\_total}}$
\State $ent_i = \frac{c\_val[i]}{c\_total}\sum\limits_{j}{-\frac{c\_bins[i][j]}{c\_val[i]}*\log\frac{c\_bins[i][j]}{c\_val[i]}}$
\State $con\_ent = \sum\limits_{i}{ent_i}$
\State sort $con\_ent$ of each filter in layer $l$ ascending
\State add corresponding filter of top-$r$ $con\_ent \to \sigma^r$
\EndProcedure
\end{algorithmic}
\end{algorithm}

\subsection{Statistical result of CIFAR10 on VGG-16}
As one of the core concepts in the convolutional network, feature maps reveals huge amount of information about the information flow within the network.
For which we proposed using statistical variable, especially conditional entropy, to measure the connection between feature map and error probability, which may help us understand how the information flows through the network.

Figure~\ref{fig:3_stat_vgg} is the statistical result of filters in each convolutional layer of CIFAR10 on VGG-16.
The x-axis indicate the index of convolution layer in model.
Figure~\ref{maxv}, Figure~\ref{maxp} shows the maximum $ent_i$ and its corresponding output activation, respectively.

Figure~\ref{act_zero} depicts the number of zero activations and Figure~\ref{act_zero_rate} shows the negative zero activation ratio.
We observe zero activation mostly reside in first and last several convolutional layers.

Figure~\ref{ent} demonstrate total conditional entropy of each filter.
Figure~\ref{maxv_rate} is the ratio of maximum conditional entropy to total conditional entropy and it's remains relatively uniform in different layers.

\begin{figure}[!ht]
\centering
\begin{subfigure}{1.0\columnwidth}
	\includegraphics[width=\columnwidth]{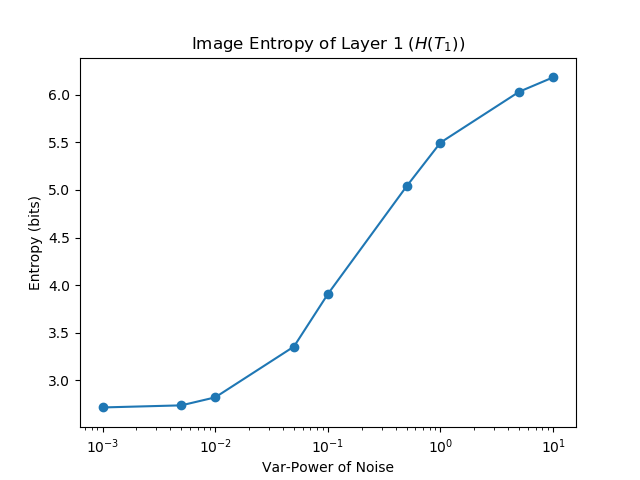}%
	\caption{First convolutional layer entropy.}%
	\label{mnist_1}%
\end{subfigure}\hfill%
\begin{subfigure}{1.0\columnwidth}
	\includegraphics[width=\columnwidth]{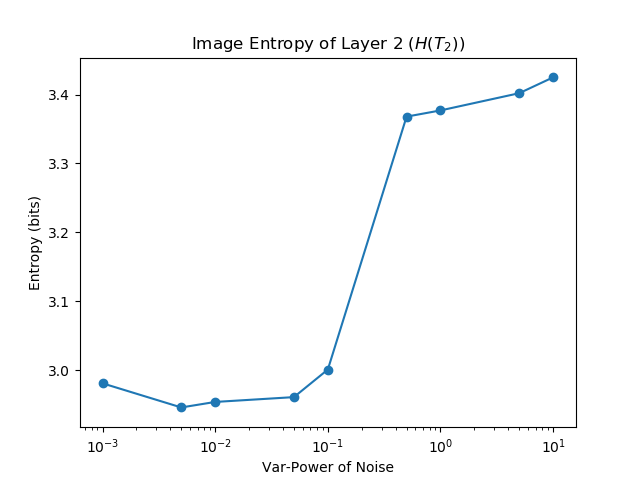}%
	\caption{Second convolutional layer entropy.}%
	\label{mnist_2}%
\end{subfigure}\hfill%
\begin{subfigure}{0.8\columnwidth}
	\includegraphics[width=\columnwidth]{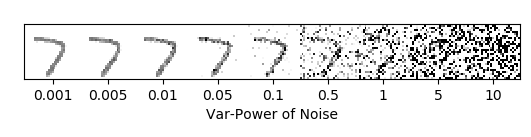}%
	\caption{Sample under different noise level.}%
	\label{mnist_noise}%
\end{subfigure}
\caption{Two layer CNN on MNIST}
\label{mnist}
\end{figure}

\section{Experiments}
\label{sec:evaluation}

\subsection{Experiment setup}
In the initial stage of our experiment, we assume using entropy $H(T_n)$ to measure information flowed in the neural network. 
To testify our assumption, we made several experiments with MNIST datasets and a two-layered CNN. 
We added Gaussian noise to the picture and see how the entropy in neural network would change as the power of noise changes. 
If the entropy effectively measures the information uncertainty in CNN, $H_{T_1}$ and $H_{T_2}$ should increase as the loss increases.

The result is affirmative, $H_{T_1}$ and $H_{T_2}$ do increase as expected. 
Yet there are still some problems about $H_{T_2}$. 
As shown in Figure~\ref{mnist}, $H_{T_2}$ does not increase as much as $H_{T_1}$ when var is below 0.1.
We assume that CNN reduces the influence of noise in the classification task when the noise is not so significant.
So the entropy does not increase so much in the second layer.

Here, we trained a VGG-16~\cite{simonyan2014very} network with \emph{tanh} activation function on CIFAR10 dataset~\cite{krizhevsky2009learning} and logged the output of each layers. 
The network has 13 convolution layers, each layer has 64 filters to 512 filters, with the size of 3.
The network converged at around 100 epochs, but we trained it for 400 epochs. 
The final accuracy is $99.6\%$ on training set and $92.68\%$ on test set. 
As we can observe from Figure~\ref{ent}, conditional entropy in filters are almost uniformly distributed, except for the first convolutional layer where conditional entropy is slightly higher.
Also, number of zero activations shown in Figure~\ref{act_zero} indicates the same fact that the first convolutional layer in VGG-16 is greatly redundant in CIFAR10 classification.

However, the philosophy that each neuron in network is interchangeable and informative equivalent~\cite{williamson2015equivalence} indicate a uniform distribution of information across the layers.
As a result, filters are supposed to be pruned in a layer-wise manner where the number of filters to be pruned should be proportional in each layer, e.g. prune 16 filters in convolutional layer with 64 filters, 32 filters with 128 etc.

In the experiment, we evaluate trade-off between accuracy and pruning ratio in global and layer-wise approach respectively.
Inspired from above observation, we propose a two phase filter pruning framework based on conditional entropy, namely \textit{2PFPCE}.
The filter selection criteria is described in Algorithm~\ref{algo:3_procedure}.
The procedure is shown in Figure~\ref{fig:4_exp_method}.

In phase I, filters are pruned and fine-tuned iteratively until the accuracy drop reach the threshold (1\%).
The purpose is to remove the redundancy filters with respect to the dataset.
Notice that dataset plays a crucial role when it comes to compression.
The number of involved features in a 1000-category dataset is probably much larger than that in a 10-category dataset.
After a layer is pruned, weights in the filter w.r.t the maximum activation entropy are kept constant and cannot be updated during fine tuning.
Similar to dropout~\cite{srivastava2014dropout}, this aims at penalizing any single neuron that may overly fitted to the dataset.

In phase II, filters are pruned in a layer-wise manner: in each iteration, a small portion filters (1/32 or 1/16) of each layer are pruned until the accuracy drop reach the threshold $\gamma$.
The threshold $\gamma$ is a hyper-parameter and can be adjusted to satisfy application constraint.
To retain the information in pruned filters and avoid time consuming fine-tune, we update the bias term as the activation of maximum conditional entropy filter (Figure~\ref{maxp}).


\subsection{Global Pruning Approach}
\label{sec:exp_glob}

\begin{figure}[t]
\centering
\includegraphics[width=1.0\columnwidth, trim={3cm 1.6cm 3cm 2.5cm},clip]{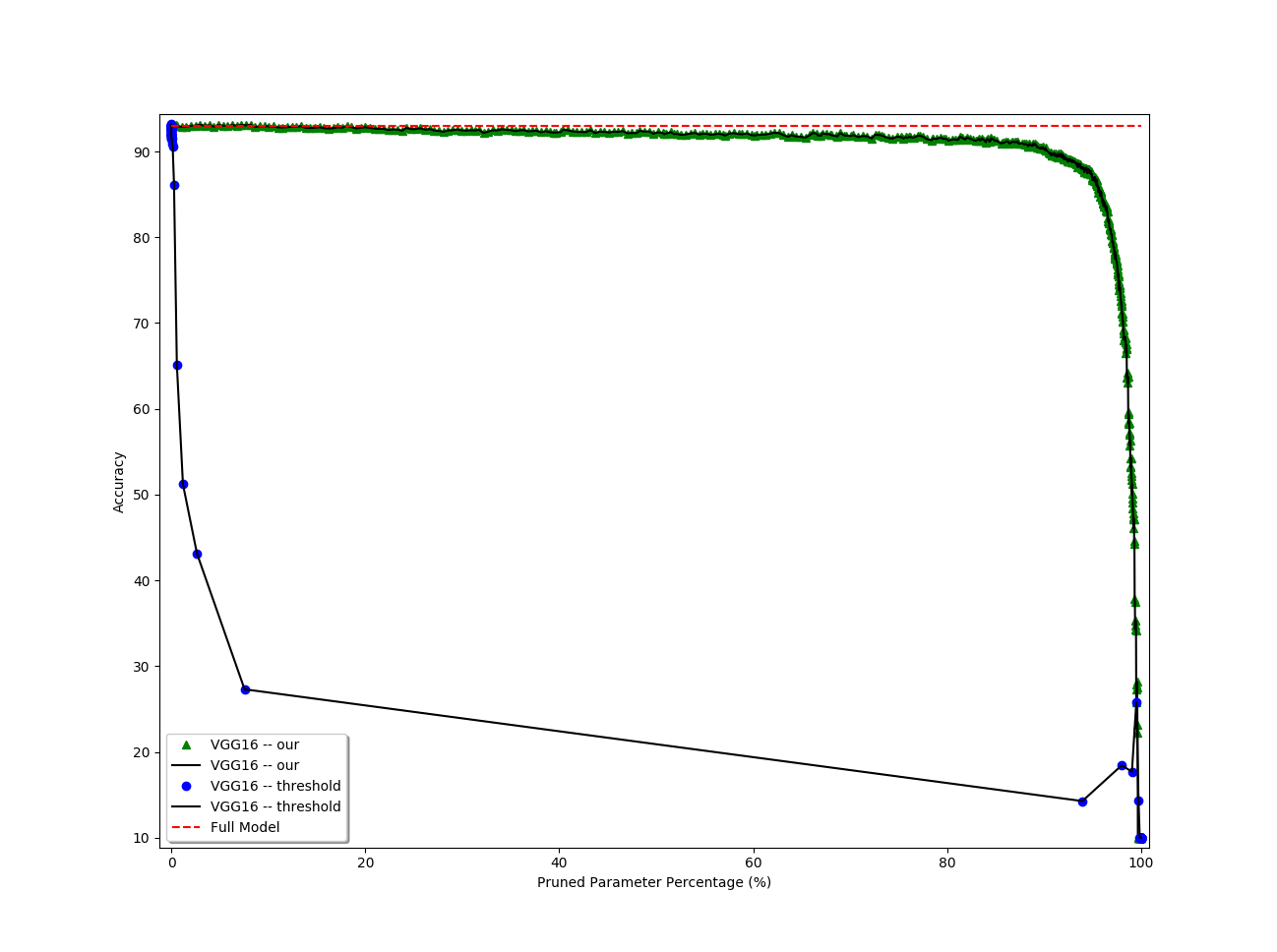}
\caption{Accuracy vs. pruning ratio of VGG16.}
\label{fig:4_glob_vgg}
\end{figure}

\begin{figure}[t]
\centering
\includegraphics[width=0.98\columnwidth, trim={3cm 1.5cm 2.7cm 2.8cm},clip]{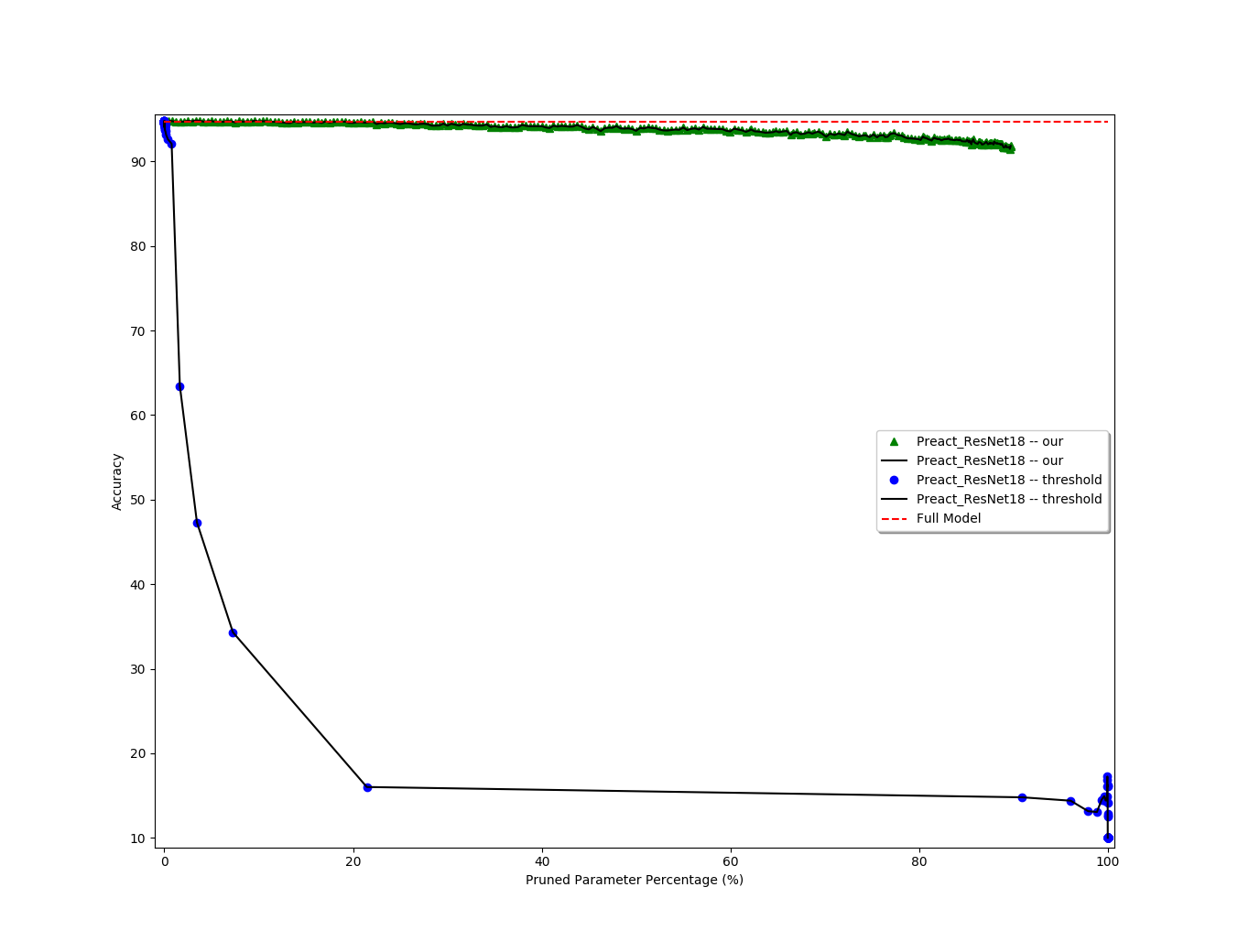}
\caption{Accuracy vs. pruning ratio of Preact-ResNet18.}
\vspace{-6pt}
\label{fig:4_glob_resnet}
\end{figure}

In this experiment, we evaluate the trade-offs between computation complexity and classification accuracy in global pruning approach.
As shown in Figure~\ref{fig:4_glob_vgg} and Figure~\ref{fig:4_glob_resnet}, the accuracy w.r.t pruning ratio is similar in both networks.
Without fine-tuning, the accuracy in VGG-16 and ResNet-18 decrease significantly to 27\% and 16\% within a pruning ratio of 10\% and 21\%, respectively.
After the turning point, in VGG-16 the accuracy slowly decrease to 15\% as the pruning ratio increase while in ResNet-18 the accuracy drop is within 1\%.
With fine-tuning, each network can remain its accuracy with pruning ratio below 20\% and then slowly degrade with a marginal accuracy drop between 1\%-2\% until the pruning ratio reaches 90\%.
The comparison between w/o fine-tune shows the DNNs can recover from small disturb with fine tuning.
The contribution of each filter is very similar in a converged convolutional neural network.
Though the experiment results on VGG-16 and ResNet-18, we can conclude that our current solution can prune about 40\% filters with 1\% accuracy compromise and 80\% prunes with only 2\% accuracy compromise with global pruning approach.

\subsection{Layer-wise Pruning Approach}
\label{sec:exp_layer}

Table~\ref{table:res_cifar10_1} compares the obtained accuracy of different filter importance criterion under the same pruning ratios.
For a fair comparison, the DNNs are pruned in a layer-wise approach and fine-tuned same number of epochs for accuracy recover after pruning.
At each pruning iteration, we remove a percentage of feature maps and then perform 20 minibatch SGD updates with batch-size 32, momentum 0.9, learning rate $10^{-4}$, and weight decay $10^{-4}$. 
As we can see from the table, our proposed conditional entropy filter selection criteria outperforms counterparts with highest accuracy on both pruning ratio 25\% and 50\%.

Table~\ref{table:res_cifar10_2} compares the obtained pruning ratio of filter importance criterion under approximately the same errors, e.g., within 2\% accuracy loss.
The baseline is L1-Norm which has a pruning ratio of 60\% on both networks.
Our experiments show that with a marginal accuracy loss, conditional entropy based criterion can always achieve largest pruning ratio. 

\begin{table}
\centering
\begin{subtable}{0.5\textwidth}
\centering
\caption{Filter importance criteria vs. accuracy.}
\label{table:res_cifar10_1}
\vspace{4pt}
\begin{tabular}{ |p{1.0cm}|p{1.4cm}|p{1.4cm}|p{1.4cm}|p{1.4cm}| }
	\hline
    \multicolumn{5}{|c|}{VGG16/CIFAR10: 92.98\%} \\
    \hline
    Prune & L1-Norm & APoZ & Act. ent & Cond. ent\\
    \hline
    25\% & 92.86\% & 92.94\% & 92.93\% & 93.50\% \\
    50\% & 92.11\% & 92.02\% & 92.00\% & 92.76\% \\
    \hline
    \multicolumn{5}{|c|}{ResNet50/CIFAR10: 93.16\% } \\
    \hline
    20\% & 94.42\% & 94.36\% & 94.33\% & 94.84\% \\
    50\% & 94.48\% & 94.25\% & 94.42\% & 94.44\% \\
    \hline
\end{tabular}
\end{subtable}
\begin{subtable}{0.5\textwidth}
\centering
\vspace{6pt}
\caption{Filter importance criteria vs. pruning ratio.}
\label{table:res_cifar10_2}
\vspace{4pt}
\begin{tabular}{ |p{1.6cm}|p{1.35cm}|p{1.1cm}|p{1.35cm}|p{1.35cm}| }
	\hline
    \multicolumn{5}{|c|}{VGG16/CIFAR10: 92.98\%} \\
    \hline
    Acc. & L1-Norm & APoZ & Act. ent & Cond. ent\\
    \hline
    91.0($\pm$0.3)\% & $1.0\times$ & $0.88\times$ & $1.27\times$ & $1.32\times$ \\
    \hline
    \multicolumn{5}{|c|}{ResNet50/CIFAR10: 93.16\% } \\
    \hline
    92.0($\pm$0.2)\% & $1.0\times$ & $0.93\times$ & $0.96\times$ & $1.05\times$ \\
    \hline
\end{tabular}
\end{subtable}
\caption{Comparison of filter importance criteria in layer-wise approach VGG-16/ResNet-50 on CIFAR10}
\label{4_layer_cmp}
\end{table}

\subsection{Stage Pruning Approach}

\begin{figure}[t]
\centering
\includegraphics[width=1.0\columnwidth]{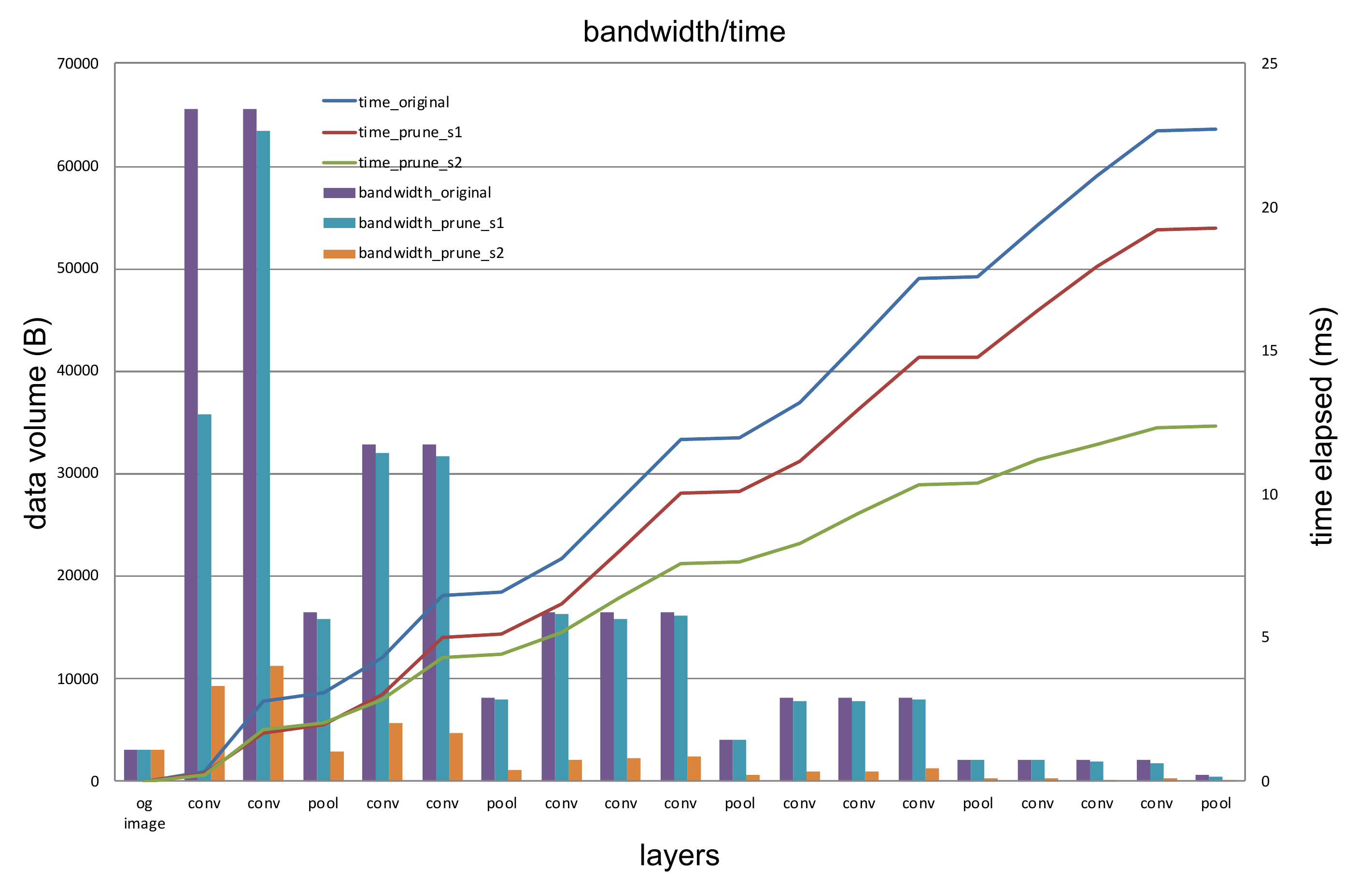}
\caption{Stage pruning with 98\% accuracy threshold.}
\label{fig:4_stage_all}
\end{figure}

We focus on reducing the number of convolutional feature maps and the total estimated floating point operations (FLOPs).
During pruning we were measuring reduction in computations by FLOPs, which is a common practice~\cite{han2015deep}. 
Improvements in FLOPs result in monotonically decreasing inference time of the networks because of removing entire feature map from the layer. 
However, time consumed by inference dependents on particular implementation of convolution operator, parallelization algorithm, hardware, scheduling, memory transfer rate etc.
Unlike in previous work, where latency is considered via another, often
inaccurate proxy (e.g., FLOPS), in our experiments, we directly measure real-world inference time by executing the model on GPU (NVIDIA Titan Xp with CUDNN 8.0).
Therefore we measure improvement in the inference time to see real speed up compared to original networks in Figure~\ref{fig:4_stage_all}. 

Based on the observations mentioned in setup, a two-phase pruning approach is proposed.
As shown in Figure~\ref{fig:4_stage_all}, a combination of both approach can achieve a pruning ratio of 88\% within 2\% accuracy drop.
The dataset is CIFAR10 with a minibatch size of 32.
The inference time on pre-trained VGG-16 is 22.69ms.
In Phase I, number of filters decrease from 4224 to 3960 and total bandwidth from 310784 bytes to 273658 bytes.
The inference time reduces to 19.24ms.
Removed filters are mostly from the first convolutional layer and this verifies our observation on redundancy.
In the first convolutional layer, $\sim 60\%$ filters are removed result in a reduction in data volume from 64M to 35M.
Because of high parallelization in GPU, inference time decrease in a single layer is not as significant as in data volume.
In Phase II, number of filters decrease from 3960 to 532 and total bandwidth from 273658 bytes to 49165 bytes.
The inference time reduces to 12.34ms.

\section{Conclusion}
\label{sec:conclusion}
In this work, inspired from the statistical result of information flow in neural network, we propose to use conditional entropy as the filter selection criteria in filter pruning.
The performance of conditional based filter selection criteria outperforms approaches based L1-Norm, APoZ and activation entropy.
Experimental result shows the proposed criteria can achieve 92.76\% accuracy when pruning ratio is 50\% and $\sim 91\%$ accuracy when pruning ratio is 80\%.
In both VGG-16 and ResNet-50, our proposed conditional entropy outperforms the above criterion.
To comply with the network information distribution, we adopt a two phase pruning framework which combines global approach with layer-wise approach.
In addition, novel model tuning techniques are proposed 1) freeze weights w.r.t maximum activation entropy to avoid over-fitting, 2) update bias w.r.t the activation of the maximum conditional entropy filter.
The above framework can achieve a pruning ratio of 88\% within 2\% accuracy drop of pre-trained VGG-16 model on CIFAR10.

\bibliographystyle{plain}
\bibliography{aaai.bib}
\end{document}